# FHHOP: A Factored Hybrid Heuristic Online Planning Algorithm for Large POMDPs

**Zongzhang Zhang**
School of Computer Science and Technology
University of Science and Technology of China
Hefei, Anhui 230027 China
zzz@mail.ustc.edu.cn

**Xiaoping Chen**
School of Computer Science and Technology
University of Science and Technology of China
Hefei, Anhui 230027 China
xpchen@ustc.edu.cn

## Abstract

Planning in partially observable Markov decision processes (POMDPs) remains a challenging topic in the artificial intelligence community, in spite of recent impressive progress in approximation techniques. Previous research has indicated that online planning approaches are promising in handling large-scale POMDP domains efficiently as they make decisions "on demand" instead of proactively for the entire state space. We present a Factored Hybrid Heuristic Online Planning (FHHOP) algorithm for large POMDPs. FHHOP gets its power by combining a novel hybrid heuristic search strategy with a recently developed factored state representation. On several benchmark problems, FHHOP substantially outperformed state-of-the-art online heuristic search approaches in terms of both scalability and quality.

## 1 Introduction

Partially observable Markov decision processes (POMDPs) have been widely recognized as a powerful probabilistic model for planning and control problems in partially observable stochastic domains (Kaelbling et al. 1998). Solving POMDP problems exactly can be impossible due to their computational complexity (Madani et al. 1999). In the past decade, researchers have made impressive progress in designing approximate algorithms (Pineau et al. 2006; Kurniawati et al. 2008; Ross et al. 2008a) and have successfully applied them to various realistic robotic tasks (Hoey et al. 2007; Hsu et al. 2008).

Planning algorithms can be categorized as offline planning algorithms and online planning algorithms depending on the ways of solving problems. Offline approaches separate the policy-construction phase and the policy-execution phase. They devote significant pre-processing time to generate a policy over the whole belief space in the policy-construction phase, then use the resulting policy to make decisions during the policy-execution phase. Offline approaches can be beneficial for repeated POMDP planning tasks as the pre-processing time can be amortized over the various runs. In contrast, online approaches are a viable alternative for urgent or one-off POMDP planning tasks. They do not allocate significant time to pre-processing, but alternate between a time-limited policy-construction step for the current state and a policy-execution step (Ross et al. 2008a). Instead of finding policies that generalize to all possible situations, they focus on computing good local policies at each policy-construction step, and therefore can potentially produce a sequence of actions with high reward while spending much less overall time for policy construction and policy execution compared to offline algorithms.

In this paper, we aim to accelerate online POMDP planning algorithms by taking advantage of a more efficient heuristic search strategy and a factored state representation.

Our work on heuristics originates from the insight that lower bounds on the optimal value function have not been well exploited in current online algorithms. Current online algorithms maintain both the lower and upper bounds as a heuristic to find good policies. One key problem that determines the overall performance of current approaches is how to approximate the optimal policy according to the lower and upper bounds. In previous work (Ross et al. 2008a), the upper bound was usually more popular than the lower bound in the calculation. The search toward the action with the highest upper bound guarantees that an $\epsilon$-optimal action can be found within finite time theoretically. In contrast, the lower bound is more difficult to be exploited since it can easily trap search into local optima. For example, the result of always exploring toward the action branch with the highest lower bound only makes us believe that this action branch is optimal, although it tends not to be so. However, the lower bound can guarantee the quality of policy, which is an advantage that the upper bound does not have. Thus, online algorithms typically return the best action with the highest lower bound but not with the highest upper bound. In this paper, we discuss how to take full advantage of both the lower and upper bound in a novel hybrid way and avoid their disadvantages. The role of the lower bound in our heuristic method is to guide the search toward a set of promising policies whose quality may be better than the current best policy.

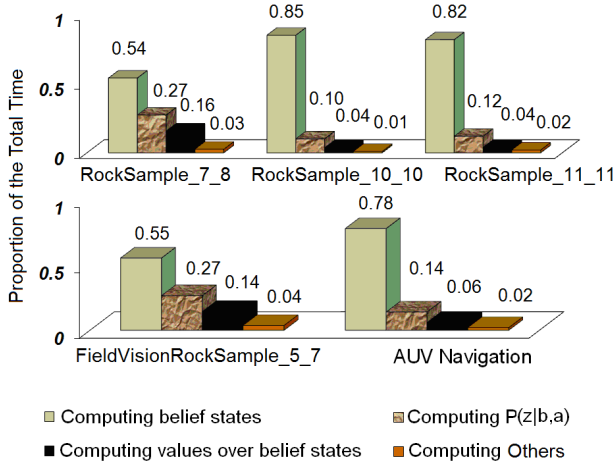

Figure 1: AEMS2's computational profile on several benchmark problems. The plot shows, for example, that on RockSample_7_8 AEMS2 spends 54%, 27% and 16% of its running time in computing new belief states, $P(z|b,a)$, and values over belief states given a value function, respectively. See text for further explanations.

Our work on factorization was inspired by profiling data on several typical benchmark problems like those in Figure 1. Ignoring the details of the tasks for the time being, we can see that in all cases the anytime error minimization search 2 (AEMS2) algorithm, one of the most efficient online heuristic search algorithms (Ross and Chaib-draa 2007; Ross et al. 2008a), spends more than 95% of its overall running time in computing: (1) new belief states; (2) $P(z|b,a)$, the probabilities of observing $z$ after taking action $a$ at belief state $b$; and (3) values over belief states given a value function. These dramatic observations drive us to decrease overall computation time by reducing the time on these three operations. We show theoretically and empirically that it is possible to do so by taking advantage of the mixed observability MDP (MOMDP) representation (Ong et al. 2010). The MOMDP representation can be considered as an instance of a dynamic Bayesian network (DBN). It exploits mixed observability where some state variables are always observed and others are hidden to reduce the time complexity of state related operations.

We refer to our new algorithm based on the above two insights as Factored Hybrid Heuristic Online Planning (FHHOP). Experimental results reveal that FHHOP substantially outperformed the AEMS2 algorithm on all test problems. Especially, on some well-known benchmark problems, FHHOP has achieved more than an order of magnitude improvement in terms of runtime.

The remainder of the paper is organized as follows. Section 2 provides an overview of the POMDP model, online algorithms and the MOMDP representation. Section 3 describes the details in the FHHOP algorithm, which is our key contribution. Section 4 describes a set of experiments showing the efficacy of the FHHOP algorithm in terms of both runtime and solution quality. We conclude and discuss future work in Section 5.

## 2 Background and Related Work

In this section, we briefly introduce the POMDP model, online algorithms, and the MOMDP representation.

### 2.1 POMDP Model

POMDPs provide a very powerful mathematical model for an agent's decision making in partially observable domains. A discrete and discounted POMDP model can be formally defined by a tuple $(S, A, Z, T, \Omega, R, \gamma)$. In the tuple, $S$, $A$ and $Z$ are the finite and discrete state space, action space and observation space, respectively, $T(s, a, s') : S \times A \times S \rightarrow [0, 1]$ is the state transition function $(P(s'|s, a))$, $\Omega(a, s', z) : A \times S \times Z \rightarrow [0, 1]$ is the observation function $(P(z|a, s'))$, $R(s, a) : S \times A \rightarrow \mathbb{R}$ is the reward function, and $\gamma \in (0, 1)$ is the discount factor on the summed sequence of rewards. Because the agent's current state is not fully observable, the agent could have to rely on the complete history of past actions and observations to select a desirable current action. In the context of decision making, the *belief state* $b$ is a sufficient statistic for the history of actions and observations (Smallwood and Sondik 1973). A belief state $b$ is a discrete probability distribution over the state space, whose element $b(s)$ gives the probability that the agent's state is $s$. Let $\mathcal{B}$ be the space of all possible belief states and $b_0$ be the agent's *initial belief state*. Thus, $\mathcal{B}$ is an $|S|$-dimensional space, where $|S|$ is the number of states. In a so-called flat POMDP, all operations on beliefs and value functions are performed at the level of $|S|$-dimensional belief states.

When the agent takes action $a$ at a belief state $b$ and receives observation $z$, it will arrive at a new belief state $b^{a,z}$:

$$b^{a,z}(s') = \frac{1}{\eta} \Omega(a, s', z) \sum_{s \in S} T(s, a, s') b(s), \quad (1)$$

where $\eta$ is a normalizing constant (Kaelbling et al. 1998). The constant is the probability of receiving $z$ after the agent

takes $a$ at $b$ and can be specified as:

$$P(z|b,a) = \sum_{s' \in S} \Omega(a, s', z) \sum_{s \in S} T(s, a, s')b(s). \quad (2)$$

A key goal in solving POMDPs is to find an *optimal policy* $\pi^*$ that maximizes the *expected discounted reward* $V^\pi(b_0) = E[\sum_{t=0}^{\infty} \gamma^t R(s_t, a_t)|b_0, \pi]$, where $\pi$ denotes a policy that maps from belief states to actions, and $s_t$ and $a_t$ are the agent's state and action at time step $t$, respectively. The maximal expected discounted reward starting from any initial belief state in the whole belief space is captured by the optimal value function $V^*$, which can be defined via the fixed point of Bellman's equation (Bellman 1957):

$$V^*(b) = \max_{a \in A} Q^*(b, a), \quad (3)$$

where $Q^*(b,a) = R(b,a) + \gamma \sum_{z \in Z} P(z|b,a)V^*(b^{a,z})$ and $R(b,a) = \sum_{s \in S} R(s,a)b(s)$. The function $V^*$ can be approximated infinitely closely by a piecewise linear and convex function:

$$V(b) = \max_{\alpha \in \Gamma}(\alpha \cdot b) = \max_{\alpha \in \Gamma} \sum_{s \in S} \alpha(s)b(s), \quad (4)$$

where $\Gamma$ is a finite set of $|S|$-dimensional hyperplanes, called $\alpha$-vectors, over $\mathcal{B}$ (Smallwood and Sondik 1973). The piecewise linear and convex property makes several exact and approximate value iteration algorithms feasible for solving POMDPs (Kaelbling et al. 1998; Kurniawati et al. 2008). Once $V^*$ has been identified, computing the optimal policy $\pi^*$ is a straightforward application of $\pi^*(b) = \operatorname{argmax}_{a \in A} Q^*(b, a)$.

### 2.2 Approximate Online Algorithms

Online POMDP algorithms can be classified into three categories: branch-and-bound pruning, Monte-Carlo sampling and heuristic search (Ross et al. 2008a). In this paper, we only focus on one category, heuristic search online algorithms. The central idea in heuristic search online algorithms is to maintain lower and upper bounds on the value function, $\underline{V}(b) \leq V^*(b) \leq \bar{V}(b)$, and to iteratively improve these bounds at the current belief state $b_c$ by expanding an AND/OR tree of reachable belief states from $b_c$. We denote $\underline{Q}(b,a)$ and $\bar{Q}(b,a)$, respectively, as the lower and upper bounds over $Q^*(b,a)$. Each OR-node in the tree represents a belief state and each AND-node represents an action choice from the belief state (node) above it, given the tree is drawn growing downwards. By abuse of notation, we will use $b$ to represent a belief node in the tree and its associated belief state.

Algorithm 1 is a generic POMDP solver. It accepts three parameters: $b_0$, $\tau$ and $\epsilon$, where $b_0$ is the initial belief state, $\tau$ is a bound on computation time allowed per policy-construction step, and $\epsilon$ is the desired precision on $V(b_c)$.

---

**Algorithm 1:** Generic Online POMDP Solver

**Function** OnlinePOMDPAlgorithm($b_0$,$\tau$,$\epsilon$)
1: Initialize the bounds by offline algorithms;
2: $b_c = b_0$;
3: Build an AND/OR tree to contain $b_c$ at the root;
4: **while** $b_c$ is not a goal state
5:     $a$ =Search($b_c$,$\tau$,$\epsilon$);
6:     Take action $a$ and receive a new observation $z$;
7:     $b_c = b_c^{a,z}$;
8:     Update the tree so that $b_c$ is the new root;
9: **end while**

---

**Algorithm 2:** Policy Construction

**Function** Search($b_c$,$\tau$,$\epsilon$)
1: StartTimer();
2: **while** ElapsedTime() $\leq \tau$ and $\bar{V}(b_c) - \underline{V}(b_c) > \epsilon$
3:     $b^*$ =ChooseBestNodetoExpand();
4:     Expand($b^*$);
5:     UpdateAncestors($b^*$);
6: **end while**
7: return argmax$_{a' \in A} \underline{Q}(b_c, a')$;

---

Online algorithms use this tree to alternate between policy construction and policy execution. The policy execution procedure (see Lines 6–8 in Algorithm 1) is consistent across all online algorithms.

Algorithm 2 is a general online policy-construction function. The ChooseBestNodetoExpand() procedure uses efficient heuristic search strategies to select a useful belief node to expand among the leaf nodes. The Expand() procedure expands a belief node, improving its lower and upper bound over $V^*$ in the process. The UpdateAncestors() procedure concentrates on updating the bounds of a belief node's ancestors. On Line 7 of Algorithm 2, it returns the action branch at the root with the highest lower bound of $\underline{Q}(b_c, a')$. We define the current best policy $\pi_{best}$ to be the policy of always selecting the action $\pi_{best}(b) = \operatorname{argmax}_{a' \in A} \underline{Q}(b, a')$.

To better describe our work on heuristics, we first use Figure 2 as an example to further illustrate how online algorithms work. Suppose that $b_{c+k}$ is the leaf belief node that ChooseBestNodetoExpand() returns. Let $b_{c+k}$ be a leaf node that is reachable from the root belief node $b_c$ by following a $k$-step action-observation sequence, $a_0z_1a_1z_2 \ldots a_{k-1}z_k$. After expanding $b_{c+k}$, an improved $\underline{V}(b_{c+k})$ and $\bar{V}(b_{c+k})$ can be obtained. Furthermore, the invocation of UpdateAncestors() results in the improved bounds at all belief nodes and $Q$ nodes along this path. In other words, given that $b_{c+k}$ is reachable from $b_c$ by following some policy $\pi$, the execution of Line 3–5 in Algorithm 2 will lead to improved $\underline{V}^\pi(b_c)$ and $\bar{V}^\pi(b_c)$, the lower and upper bounds on the expected discounted reward generated by following $\pi$ at $b_c$. If $\underline{V}^{\pi_{best}}(b_c)$ in the previous round is surpassed by the new improved $\underline{V}^{\pi}(b_c)$, then a policy with higher lower bound will be found.

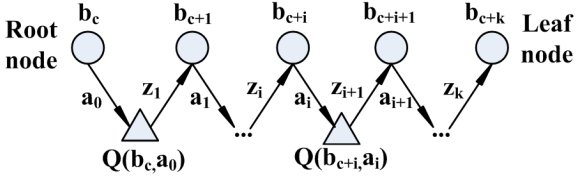

Figure 2: A path from the root belief node $b_c$ to the leaf belief node $b_{c+k}$.

### 2.3 Heuristics in Current Online Algorithms

In current heuristic search online algorithms, each leaf belief node has an associated heuristic value. The invocation of ChooseBestNodetoExpand() returns the leaf belief node with the maximal heuristic value. If an oracle provides us with the optimal value function $V^*$, a heuristic function over a leaf node $b_{c+k}$ with good theoretical guarantees can be defined as follows (Ross et al. 2008b):

$$H^*(b_{c+k}) = e^*(b_{c+k}) \prod_{t=0}^{k-1} \omega^*(b_{c+t}, a_t)\omega(b_{c+t}, a_t, z_{t+1}),$$
(5)

where $e^*(b_{c+k}) = V^*(b_{c+k}) - \underline{V}(b_{c+k})$, $\omega(b_{c+t}, a_t, z_{t+1}) = \gamma P(z_{t+1}|b_{c+t}, a_t)$, and

$$\omega^*(b_{c+t}, a_t) = \begin{cases} 1 & \text{if } a_t \in \operatorname{argmax}_{a' \in A} Q^*(b_{c+t}, a'), \\ 0 & \text{otherwise.} \end{cases}$$
(6)

In such a heuristic function, each quantity plays an important role in expanding the leaf nodes: $e^*(b_{c+k})$ encourages exploration of leaf nodes with loose bounds, $\omega^*(b_{c+t}, a_t)$ focuses the exploration toward the leaf nodes that are reachable from $b_c$ under the optimal policy, and $\omega(b_{c+t}, a_t, z_{t+1})$ guides the search toward belief nodes that are likely to be encountered in the future. Empirical results in prior work (Ross et al. 2008a) teach us that ignoring some quantities in Equation 5 would damage the overall performance of online approaches. For example, the lack of $\omega(b_{c+t}, a_t, z_{t+1})$ in the BI-POMDP algorithm (Washington 1997) may explain its poor performance on some POMDP problems with large observation spaces.

However, the optimal policy or value function is not available for constructing $H^*$. Thus, the finding of a set of "promising" policies is one of the core issues for improving the speed of online algorithms. The approach of Satia and Lave (Satia and Lave 1973) assumes the set of promising policies are all possible optimal policies. In this approach, little heuristic information about $\underline{V}$ and $\bar{V}$ is exploited to focus its search toward actions that look promising, and therefore the approach of Satia and Lave is not able to scale well to large-scale POMDPs. The AEMS1 algorithm (Ross et al. 2008a) takes both $\underline{V}$ and $\bar{V}$ into account by favoring exploration of action branches with high average values of $\underline{Q}(b, a)$ and $\bar{Q}(b, a)$. Such an algorithm loses its competitive capability in large-scale POMDP problems partially because its search strategy does not use the highest lower or upper bound well. In contrast to AEMS1, the BI-POMDP and AEMS2 algorithms use $\bar{V}$ as an estimated substitute for $V^*$. $H_U(b_{c+k})$ is the AEMS2 algorithm's heuristic:

$$H_U(b_{c+k}) = e(b_{c+k}) \prod_{t=0}^{k-1} \omega(b_{c+t}, a_t)\omega(b_{c+t}, a_t, z_{t+1}),$$
(7)

where $e(b_{c+k}) = \bar{V}(b_{c+k}) - \underline{V}(b_{c+k})$ and

$$\omega(b_{c+t}, a_t) = \begin{cases} 1 & \text{if } a_t \in \operatorname{argmax}_{a' \in A} \bar{Q}(b_{c+t}, a'), \\ 0 & \text{otherwise.} \end{cases}$$
(8)

From the definition of $\omega(b_{c+t}, a_t)$, we can see AEMS2 assumes that the action with the maximal upper bound is in fact the optimal action. Such a definition, sometimes called the IE-MAX heuristic (Smith and Simmons 2004), leads AEMS2 to find an $\epsilon$-optimal action within finite time (Ross et al. 2008b). The action branch according to the highest lower bound is deliberately ignored in AEMS2 simply because choosing leaf nodes under such an action branch to expand could only cause its lower bound to rise, and therefore, easily trap search in local optima. However, $\underline{Q}(b, a)$ may be a very useful heuristic in finding a better policy. A direct insight is that the action that Algorithm 1 returns is the action corresponding to the maximal $\underline{Q}(b, a)$ but not the maximal $\bar{Q}(b, a)$. One of our contributions is to suggest that using the lower bound could provide improvements compared to the heuristics used in AEMS2.

### 2.4 MOMDPs

A MOMDP can be generally specified as a tuple $(\mathcal{X}, \mathcal{Y}, A, Z, T_\mathcal{X}, T_\mathcal{Y}, \Omega, R, \gamma)$, where $\mathcal{X}$ is the set of fully observable state variables, $\mathcal{Y}$ is the set of partially observable state variables, $S = \mathcal{X} \times \mathcal{Y}$, $T_\mathcal{X}(x, y, a, x') : \mathcal{X} \times \mathcal{Y} \times A \times \mathcal{X} \to [0, 1]$ and $T_\mathcal{Y}(x, y, a, x', y') : \mathcal{X} \times \mathcal{Y} \times A \times \mathcal{X} \times \mathcal{Y} \to [0, 1]$ are the two corresponding probabilistic state-transition functions ($P(x'|x, y, a)$ and $P(y'|x, y, a, x')$), $\Omega(a, x', y', z) : A \times X \times Y \times Z \to [0, 1]$ is the observation function ($P(z|a, x', y')$), $R(x, y, a) : \mathcal{X} \times \mathcal{Y} \times A \to \mathbb{R}$ is the reward function, and the other quantities are the same as a POMDP model's elements. Since $x$ is fully observable, a belief state $b$ on the underlying state $s = (x, y)$ can be represented as $(x, b_\mathcal{Y}(x))$. Let $\mathcal{B}_\mathcal{Y}$ be the belief space over $\mathcal{Y}$, and $\mathcal{B}_\mathcal{Y}(x) = \{(x, b_\mathcal{Y}(x))|b_\mathcal{Y}(x) \in \mathcal{B}_\mathcal{Y}\}$, then $\mathcal{B} =$

$\bigcup_{x \in \mathcal{X}} \mathcal{B}_{\mathcal{Y}}(x)$. Thus, computations involving updating beliefs and value functions can be restricted to one of $|\mathcal{Y}|$-dimensional subspaces, $\mathcal{B}_{\mathcal{Y}}(x)$. Please note that $\mathcal{B}$ has $|\mathcal{X}||\mathcal{Y}|$ dimensions, while $\mathcal{B}_{\mathcal{Y}}$ has only $|\mathcal{Y}|$ dimensions, where $|\mathcal{X}|$ and $|\mathcal{Y}|$ are the number of fully and partially observable state variables, respectively.

After taking $a$ and receiving $z$ from $(x, b_{\mathcal{Y}}(x))$, a new belief $(x^{a,z}, b_{\mathcal{Y}}(x^{a,z}))$ results. The variable $x^{a,z}$ can be directly inferred from $z$, and $b_{\mathcal{Y}}(x^{a,z})$ can be computed as follows:

$$b_{\mathcal{Y}}(x^{a,z}, y') = \frac{1}{\eta} \Omega(a, x^{a,z}, y', z) \sum_{y \in \mathcal{Y}} T_{\mathcal{X}\mathcal{Y}} b_{\mathcal{Y}}(x, y), \quad (9)$$

where $T_{\mathcal{X}\mathcal{Y}} = T_{\mathcal{X}}(x, y, a, x^{a,z}) T_{\mathcal{Y}}(x, y, a, x^{a,z}, y')$ and $\eta$, the probability of receiving $z$ after taking $a$ at $(x, b_{\mathcal{Y}}(x))$ is

$$P(z|x, b_{\mathcal{Y}}(x), a, x^{a,z}) =$$
$$\sum_{y' \in \mathcal{Y}} \Omega(a, x^{a,z}, y', z) \sum_{y \in \mathcal{Y}} T_{\mathcal{X}\mathcal{Y}} b_{\mathcal{Y}}(x, y). \quad (10)$$

For any given $x$, $V^*(x, b_{\mathcal{Y}}(x))$ can be accurately approximated by a finite set of $|\mathcal{Y}|$-dimensional vectors $\Gamma_{\mathcal{Y}}(x)$ over $\mathcal{B}_{\mathcal{Y}}(x)$:

$$V(x, b_{\mathcal{Y}}(x)) = \max_{\alpha \in \Gamma_{\mathcal{Y}}(x)} (\alpha \cdot b_{\mathcal{Y}}(x)). \quad (11)$$

From the description of MOMDPs, we can see any MDP can be written as a MOMDP and any POMDP can be written as a MOMDP. That is because the MDP states can reside in the fully observable part of the MOMDP description and the POMDP states can reside in the partially observable part of a MOMDP description. Similarly, any MOMDP can be written as a POMDP by treating all states as partially observable (Ong et al. 2010).

## 3 Factored Hybrid Heuristic Online Planning

In this section, we describe the FHHOP algorithm in detail. First, we construct a novel hybrid heuristic strategy by combining a new heuristic function using the lower bound with an existing heuristic function using the upper bound to improve the heuristic search mechanism in the ChooseBestNodetoExpand() procedure. Then, we concentrate on the reason that the MOMDP representation is useful to reduce the running time in the Expand() and UpdateAncestors() procedures. Finally, we discuss the convergence property of the FHHOP algorithm.

### 3.1 Constructing a Heuristic Function Using the Lower Bound

As mentioned in the last paragraph of Section 2.2, to find a better policy, we need to select a policy $\pi$ and expand the leaf belief nodes that are reachable from $b_c$ by following the policy $\pi$. Our goal here is to use the lower bound as a heuristic to select a set of promising policies so that the exploration toward them will lead to find an even better policy as quickly as possible. Our idea is to elicit a set of policies $\pi$ from all possible policies, whose $\underline{V}^{\pi}(b_c)$ are close to $\underline{V}^{\pi_{best}}(b_c)$, according to the lower bound.

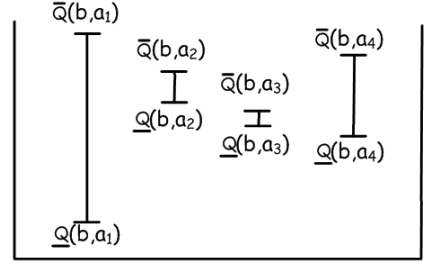

Figure 3: An example of action branches at $b$. $A_L = \{a_2\}$, $A_S = \{a_1, a_4\}$, and current second-best action is $a_4$. The action branch $a_3$ is not included in $A_S$ because $\bar{Q}(b, a_3)$ is smaller than $\underline{Q}(b, a_2)$.

First, define $\omega_1(b, a)$, a term that guides the exploration toward belief nodes that are reachable from $b_c$ under the current best policy, as follows:

$$\omega_1(b, a) = \begin{cases} 1 & \text{if } a \in \operatorname{argmax}_{a' \in A} \underline{Q}(b, a'), \\ 0 & \text{otherwise.} \end{cases} \quad (12)$$

Then, we consider the action branches at a single belief node $b$. We assume $A_L = \operatorname{argmax}_{a' \in A} \underline{Q}(b, a')$ and $A_S = \{a \in A \setminus A_L | \bar{Q}(b, a) > \max_{a' \in A} \underline{Q}(b, a')\}$. Thus, $A_L$ is a current best action branch at $b$ and $A_S$ is a set of actions without both $A_L$ and suboptimal action branches. The condition $\bar{Q}(b, a) > \max_{a' \in A} \underline{Q}(b, a')$ is used to prune suboptimal action branches at $b$, just like the branch-and-bound pruning technique in the approach of Satia and Lave. Then, we define

$$\omega_2(b, a) = \begin{cases} 1 & \text{if } a \in \operatorname{argmax}_{a' \in A_S} \underline{Q}(b, a'), \\ 0 & \text{otherwise,} \end{cases} \quad (13)$$

where $\operatorname{argmax}_{a' \in A_S} \underline{Q}(b, a')$ is called the current second-best action at $b$. So, $\omega_2(b, a)$ is a term that encourages the exploration toward belief nodes that are reachable from $b$'s second-best action branch. We use Figure 3 to provide a direct explanation of these notions. The current second-best action looks a bit like the action branch with the second highest upper bound in this figure, but they actually have nothing to do with each other.

Finally, we figure out the way of eliciting a set of promising policies according to the lower bound. Obviously, the nodes that are reachable from $b_c$ by following the set of promising policies constitute of a subtree of the current AND/OR tree. We are interested in all leaf nodes of this subtree reachable from $b_c$ through **one and only one** current second-best action branch and all other current best action branches. Such a strategy can be well depicted by $\omega_{1,2}(b_{c+k})$:

$$\omega_{1,2}(b_{c+k}) = \max_{i \in \{0,1,\ldots,k-1\}} \omega_2(b_{c+i}, a_i) \prod_{\substack{t=0 \\ t \neq i}}^{k-1} \omega_1(b_{c+t}, a_t), \quad (14)$$

where $b_{c+i}$ is the only belief node in which its current second-best action branch is chosen. When $\omega_{1,2}(b_{c+k})$ equals 1, the leaf node $b_{c+k}$ is reachable from $b_c$ by following one of such a set of promising policies. We construct a new heuristic $H_L(b)$ for each leaf node using $\omega_{1,2}(b_{c+k})$ as follows:

$$H_L(b_{c+k}) = e(b_{c+k})\omega_{1,2}(b_{c+k}) \prod_{t=0}^{k-1} \omega(b_t, a_t, z_{t+1}). \quad (15)$$

From Equation 15, we can see that the leaf nodes being selected to expand using $H_L(b)$ are not reachable from $b_c$ under the current best policy. However, they are reachable from $b_c$ under a set of promising policies that are very similar to the current best policy since they have only one action branch that is different from the current best action branch. Such a heuristic $H_L(b)$ has several advantages, as follows:

- First, a better policy according to the lower bound would be quickly found using $H_L(b)$ only if the improved $Q$-value at the second-best action branch is greater than the $Q$-value at the current best action branch.

- Second, it focuses search toward a set of promising policies but not a single policy. The main drawback of expanding the leaf nodes that are reachable under a single policy is that the time of expanding is wasted if such a single policy is not better than the current best policy. Since $H_L(b)$ selects a set of promising policies, suboptimal policies will be eliminated as the depth of its leaf nodes increases. In other words, $H_L(b)$ favors exploration toward the highest potential policy among the set of promising policies as time goes on.

- Third, such a set of promising policies is only a very small subset in the set of all possible policies. Therefore, the technique can avoid exploring a large policy space to satisfy the requirement of real-time limitations in online algorithms.

However, such a heuristic may still be at the risk of trapping the search into local optima if the optimal policy is outside of the set of promising policies and the current best policy. Thus, our new heuristic function $H_L(b)$ might be only a greedy heuristic and not lead to optimal behavior in the end.

### 3.2 Constructing a Hybrid Heuristic Strategy

Before we construct a hybrid strategy, recall the strength and weakness of $H_L(b)$ and $H_U(b)$. $H_L(b)$ is a heuristic that depends on the lower bound. Its advantage is to guide search toward finding a better policy with a higher lower bound guarantee quickly. However, $H_L(b)$ limits its search effort to a set of promising policies. If both the current policy and the set of promising policies are suboptimal, the search led by $H_L(b)$ might become trapped in a local optimum. $H_U(b)$ is a heuristic strategy that depends on the upper bound. It favors exploration toward an $\epsilon$-optimal action branch. The $\epsilon$-optimal action can be found theoretically even if $H_L(b)$ is not used. However, the process of finding an $\epsilon$-optimal action is usually very slow, especially in POMDP domains with large action and observation spaces. Because their strengths are complementary, it seems to be reasonable to combine them for obtaining a hybrid heuristic search strategy that leverages the advantages of both heuristics.

We suggest a way of constructing a hybrid strategy. Let $\mathcal{L}$ be a set of leaf belief nodes in the AND/OR tree,

$$b_U = \operatorname*{argmax}_{b \in \mathcal{L}} H_U(b) \quad (16)$$

and

$$b_L = \operatorname*{argmax}_{b \in \mathcal{L}} H_L(b). \quad (17)$$

Then, our hybrid strategy selects $b^*$ by Equation 18:

$$b^* = \begin{cases} b_U & \text{if } C_U H_U(b_U) > C_L H_L(b_L), \\ b_L & \text{otherwise,} \end{cases} \quad (18)$$

where $C_i$ denotes expected change value of both $\underline{V}(b_c)$ and $\bar{V}(b_c)$ if $b_i$ is chosen to be expanded. We use the following formula to compute $C_i$ in our experiments:

$$C_i = \frac{I_i + 1}{N_i + 1}, \quad (19)$$

where $i = U$ or $L$, $I_i$ denotes the accumulative total change value of both $\underline{V}(b_c)$ and $\bar{V}(b_c)$ caused by expanding $b_i$ for $N_i$ times, and $N_i$ represents the number of expanding $b_i$. Both $I_i$ and $N_i$ are recalculated at the beginning of each policy-construction step. The overhead of computing $C_i$ is negligible in comparison to the Expand() and UpdateAncestors() procedures. Here, $b_U$ (or $b_L$) refers to all belief nodes generated when $C_U H_U > C_L H_L$ (or otherwise), and expanding $b_i$ once means calling Expand($b_i$) and UpdateAncestors($b_i$) once. The ones appearing in both numerator and denominator are used to make $C_U$ and $C_L$ equal 1 at the beginning of the online algorithm's run. $C_U$ and $C_L$ are variables for adjusting the importance of $H_U(b)$ and $H_L(b)$, respectively, in heuristic search. For example, imagine the expansions of $b_U$ (in a period) brought no or only small change in both $\underline{V}(b_c)$ and

$\bar{V}(b_c)$. As time goes on, the value of $C_U$ will decrease, and the probability of selecting $b_U$ to expand will also drop.

Note that there are many other possible hybrid methods that combine $H_L(b)$ and $H_U(b)$. Here is only one feasible way. A further research in this direction may lead to a more efficient hybrid method.

### 3.3 Leveraging the MOMDP Representation

Now, we analyze why the MOMDP representation is a useful tool to reduce the time complexity of the operations related with belief-state computation and value-function computation. Reconsidering Equations 1, 2 and 4 in the context of POMDPs, we can see that computing $b^{a,z}$, $P(z|b,a)$ and $V(b)$ takes $|\mathcal{X}|^2|\mathcal{Y}|^2 + 2|\mathcal{X}||\mathcal{Y}|$, $|\mathcal{X}|^2|\mathcal{Y}|^2$ and $|\Gamma||\mathcal{X}||\mathcal{Y}|$ multiplications, respectively. However, by applying the MOMDP-specific versions of these equations—Equations 9, 10 and 11—these costs can be reduced to $2|\mathcal{Y}|^2 + 2|\mathcal{Y}|$, $2|\mathcal{Y}|^2$ and $|\Gamma_\mathcal{Y}(x)||\mathcal{Y}|$ multiplications, respectively. That is, the time complexity of computing $b^{a,z}$, $P(z|b,a)$ and $\underline{V}(b)$ can be reduced by $\frac{|\mathcal{X}|^2|\mathcal{Y}|^2+2|\mathcal{X}||\mathcal{Y}|}{2|\mathcal{Y}|^2+2|\mathcal{Y}|}(= \mathcal{O}(|\mathcal{X}|^2))$, $\frac{|\mathcal{X}|^2}{2}$ and $\frac{|\mathcal{X}||\Gamma|}{|\Gamma_\mathcal{Y}(x)|}$ times. Since $|\Gamma| = \sum_{x \in \mathcal{X}} |\Gamma_\mathcal{Y}(x)|$, the expectation $E_{x \sim \mathcal{X}}[\frac{|\Gamma|}{|\Gamma_\mathcal{Y}(x)|}]$ is $|\mathcal{X}|$. Thus, the MOMDP representation reduces each of the three operations' complexity by a factor of $\mathcal{O}(|\mathcal{X}|^2)$.

### 3.4 Convergence of FHHOP

The new online heuristic search algorithm using the hybrid heuristics in Section 3.2 and the POMDP representation in Section 3.3 is the FHHOP algorithm. Note that there are substantial differences between our hybrid strategy and the strategy included in AEMS1. AEMS1 merges $\underline{V}$ and $\bar{V}$ at the beginning of designing the action selection strategy in its heuristic function. However, FHHOP first separates $\underline{V}$ and $\bar{V}$ in order to define two almost independent heuristic functions, then merges the two heuristic functions with the goal of taking full advantage of each heuristic function. As a final note, Proposition 1 states the convergence property of the FHHOP algorithm.

**Proposition 1.** *Let $\epsilon > 0$ and $b_c$ the current belief state. Then, the FHHOP algorithm is guaranteed to find an $\epsilon$-optimal action for $b_c$ within finite time.*

*Proof.* Because the FHHOP algorithm exploits the MOMDP representation to accelerate the computation of $b^{a,z}$, $P(z|b,a)$ and $\underline{V}(b)$ without losing accuracy, following Theorem 2 in (Ross et al. 2008b), we only need to prove that the probability of choosing $b_U$ to expand in the FHHOP algorithm is always greater than 0 in this algorithm. The following is the proof by contradiction.

Suppose that after the $M^{th}$ expansion of $b_U$ or $b_L$, the FHHOP algorithm never selects $b_U$ to expand, namely, $C_U H_U(b_U) \leq C_L H_L(b_L)$. Thus, $I_U$ and $N_U$ will never change, and therefore, $C_U$ and $H_U(b_U)$ will never change after the $M^{th}$ expansion. Let $\underline{V}_0(b)$ and $\bar{V}_0(b)$ be $b$'s initialized lower and upper bounds, and $e_0(b) = \bar{V}_0(b) - \underline{V}_0(b)$. Then, we have

$$H_L(b_L) \leq e(b_L) \leq \max_{b \in \mathcal{B}} e_0(b),$$

and $I_L \leq e_0(b_c)$. Now, let

$$N_L > \max\{M, \frac{[e_0(b_c)+1]\max_{b \in \mathcal{B}} e_0(b)}{C_U H_U(b_U)} - 1\}.$$

Then,

$$\begin{aligned} C_U H_U(b_U) &> \frac{[e_0(b_c)+1]\max_{b \in \mathcal{B}} e_0(b)}{N_L+1} \\ &\geq \frac{I_L+1}{N_L+1} H_L(b_L) \\ &= C_L H_L(b_L). \end{aligned}$$

There is a contradiction. $\square$

## 4 Experimental Results

In this section, we present detailed empirical results designed to test the performance of the FHHOP algorithm. In all test domains, we use the blind policy and the FIB method (Hauskrecht 2000) to initialize the lower and upper bounds, respectively. Except where stated otherwise, all experiments were run on an AMD dual core processor 3600+ 2.00GHz with 2GB memory. We implemented FHHOP and AEMS2 using C++, within APPL-0.93[1], an efficient POMDP solver. Results about SARSOP without attribution appeared in Table 1 were also obtained using APPL-0.93 in our experimental platform.

### 4.1 Benchmark Problems

a) *Hallway:* The Hallway domain (Littman et al. 1995) models a robot navigating in an office environment. The task of the robot in this problem is to arrive in a single goal location. The state of the robot is comprised of its current position and its current orientation. However, neither attribute is fully observable. There is a set of 5 actions for movements:{Forward, Turn-left, Turn-right, Turn-around, Stay-in-place}. The robot can partially observe the existence of walls in four directions using four independent, short-range, and noisy sensors.

b) *Tag:* The Tag problem was first described in (Pineau et al. 2003). In this environment, a mobile robot moves in a grid map with 29 positions with the goal of tagging an opponent that intentionally moves away. Both the robot and the opponent are located initially in independently selected random positions. The robot can choose either to move into one of four adjacent positions by actions {North, South, East, West} or to tag the opponent by a "Catch" action. The effect of the robot's action is deterministic. The robot can know its current position exactly, but the opponent's position is not observable for the robot unless they are in the same position. The robot tries to catch the opponent as quickly as possible to receive a good reward since each move for the robot is expensive. In the MOMDP representation for Tag, the fully observable state variable $x$ represents the robot's position. The partially observable state variable $y$ is the opponent's position.

---
[1] The software package is available from the web-page: http://bigbird.comp.nus.edu.sg/pmwiki/farm/appl/. All test problems in the MOMDP representation conform to the syntax of pomdpx specified on the same web site.

c) *RockSample:* The RockSample domain was originally presented in (Smith and Simmons 2004) and has been frequently used in recent work to test new POMDP algorithms (Ross et al. 2008a; Bonet and Geffner 2009; Ong et al. 2010; Silver and Veness 2010). This domain models a planetary robot that has to explore an area represented as a grid map and sample rocks with a scientific value. The robot receives a positive or negative reward depending on whether the sampled rock has a scientific value. RockSample_$n$_$k$ is a family of problems in the RockSample domain with a map size $n \times n$ and $k$ rocks. In these problems, the robot's action set consists of $k + 5$ actions: {North, South, East, West, Sample, Check$_1$,... Check$_k$}. The robot always knows its own position and rock positions in the map exactly. That is rock locations are fixed and encoded in the map and need not be encoded in state variables. However, whether rock $m$ has a scientific value is only partially observed via the action Check$_m$. The Check$_m$ action is responsible for gathering information about rock $m$ using a noisy long-range sensor. The accuracy of the information gathered depends on the distance between the robot and the rock checked. In the MOMDP representation for this domain, the $x$ variable is the robot's position, and the $y$ variable is a Boolean that indicates whether a particular rock has a scientific value.

d) *FieldVisionRockSample:* The FieldVisionRockSample domain was initially introduced in the work on AEMS2 (Ross and Chaib-draa 2007). This domain can be viewed as a variant of the RockSample domain. The only difference between them is the way of perceiving the rocks in the environment. In contrast to perform a check action on a specific rock to observe its state in RockSample, in FieldVisionRockSample the states of all rocks are observed after any action by the same noisy sensor. Consequently, the robot can only perform 5 actions: {North, South, East, West, Sample}, and has an observation space with size $2^k$ for the problem with $k$ rocks. What is different from the RockSample domain is that, in the MOMDP model for this task, the $y$ variable represents $k$ Boolean values, each of which indicates whether a rock has a scientific value.

e) *AUV Navigation:* The AUV Navigation problem first appeared in (Ong et al. 2009). This problem models a 3-D oceanic environment with 4 levels and a $7 \times 20$ grid map at each level. An autonomous underwater vehicle (AUV) can achieve reward by navigating from some uncertain starting position to some goal positions and avoiding rock formations. The AUV can perform 6 actions: {Forward, Stay, Left, Right, Up, Down}. All of them are stochastic due to control uncertainty or ocean currents. The AUV is equipped with an accurate pressure senor, an accurate compass, and a global positioning system (GPS), whose signals can only be received when the AUV is located at the surface level. In the MOMDP representation, the $x$ variable models the AUV's depth and orientation, and the $y$ variable models the AUV's horizonal location.

These benchmark problems were selected to illustrate specific characteristics. First, the Tag problem has a dynamic environment that changes over time even if the robot does not take any action. Second, the RockSample domain has a larger action space but smaller observation space, while the other domains have smaller action spaces but larger observation spaces. Third, the Hallway problem does not include the mixed observation structure. Fourth, all test problems except Hallway have large state spaces ranging from about $10^3$ to $10^5$ dimensions. Thus, we hope including problems with different features helps test FHHOP's overall performance thoroughly.

### 4.2 Performance Comparison

We applied FHHOP and existing state-of-the-art online and offline algorithms to these benchmark problems. Table 1 compares them in terms of the following characteristic metrics (Ross et al. 2008a): expected discounted reward (Reward), upper bound of online time per policy-construction step ($\tau$) (seconds), offline time (seconds). Looking at the comparison between our implementation of AEMS2 and FHHOP, we can find that FHHOP obtains higher values in terms of reward on all test problems with online time less or equal to AEMS2's. Furthermore, at least on the Tag, RockSample and AUVNavigation domains, FHHOP has achieved more than an order of magnitude improvement compared to AEMS2. These empirical results suggest that FHHOP tends to have a better scalability especially in large-scale POMDP domains.

Previously published results for SARSOP, except those on the Hallway and FieldVisionRockSample domains from our experimental platform, are provided in the table for comparison. Through the comparison can be found that the FHHOP algorithm is very competitive, sometimes even better, in terms of expected discounted reward, and meanwhile have advantages such as a small initial offline planning time and a small re-planning time between action

Table 1: Multi-algorithm performance comparison on several benchmark problems. See text for explanations.

| Method | Reward | Online Time $\tau$ | Offline Time | Method | Reward | Online Time $\tau$ | Offline Time |
|---|---|---|---|---|---|---|---|
| **Hallway** $(|S|=61, |\mathcal{X}|=1, |\mathcal{Y}|=61, |A|=5, |Z|=21)$ | | | | **Tag** $(|S|=870, |\mathcal{X}|=30, |\mathcal{Y}|=29, |A|=5, |Z|=30)$ | | | |
| AEMS2 | 0.50±0.01 | 0.20 | 0.02 | AEMS2 | -6.51±0.14 | 1.00 | 0.28 |
| **FHHOP** | **0.52±0.01** | **0.20** | **0.02** | **FHHOP** | **-6.04±0.14** | **0.10** | **0.12** |
| SARSOP | 0.52±0.01 | 0.00 | 1.05 | SARSOP(Ong) | -6.03±0.12 | 0.00 | 16.50 |
| **RockSample_7_8** $(|S|=12545, |\mathcal{X}|=50, |\mathcal{Y}|=256, |A|=13, |Z|=2)$ | | | | **RockSample_11_11** $(|S|=247809, |\mathcal{X}|=122, |\mathcal{Y}|=2048, |A|=16, |Z|=2)$ | | | |
| AEMS2 | 20.66±0.29 | 1.00 | 3.34 | AEMS2 | 21.11±0.28 | 10.00 | 56.81 |
| **FHHOP** | **21.45±0.30** | **0.10** | **0.51** | **FHHOP** | **21.49±0.32** | **1.00** | **7.97** |
| POMCP(Silver) | 20.71±0.21 | 1.00 | n.v. | POMCP(Silver) | 20.01±0.23 | 1.00 | n.v. |
| SARSOP(Ong) | 21.39±0.01 | 0.00 | 810.00 | SARSOP(Ong) | 21.56±0.11 | 0.00 | 1369.00 |
| **FieldVisionRockSample_5_5** $(|S|=801, |\mathcal{X}|=26, |\mathcal{Y}|=32, |A|=5, |Z|=32)$ | | | | **FieldVisionRockSample_5_7** $(|S|=3201, |\mathcal{X}|=26, |\mathcal{Y}|=128, |A|=5, |Z|=128)$ | | | |
| AEMS2 | 21.02±0.32 | 1.00 | 0.10 | AEMS2 | 22.15±0.32 | 1.00 | 0.71 |
| **FHHOP** | **22.56±0.33** | **0.20** | **0.02** | **FHHOP** | **24.46±0.34** | **0.20** | **0.12** |
| SARSOP | 23.20±0.33 | 0.00 | 508.40 | SARSOP | 29.48±0.34 | 0.00 | 1029.38 |
| **RockSample_10_10** $(|S|=102401, |\mathcal{X}|=101, |\mathcal{Y}|=1024, |A|=15, |Z|=2)$ | | | | **AUV Navigation** $(|S|=13536, |\mathcal{X}|=96, |\mathcal{Y}|=141, |A|=6, |Z|=144)$ | | | |
| AEMS2 | 20.78±0.27 | 10.00 | 11.24 | AEMS2 | 1034.89±8.36 | 100.00 | 83.21 |
| **FHHOP** | **21.35±0.34** | **1.00** | **2.82** | **FHHOP** | **1059.22±8.49** | **10.00** | **16.02** |
| SARSOP(Ong) | 21.47±0.11 | 0.00 | 1589.00 | SARSOP(Ong) | 1019.80±9.70 | 0.00 | 409.00 |

n.v.=not available     (Ong)= (Ong et al. 2009)     (Silver)= (Silver and Veness 2010)

executions. Note that there also exists other state-of-the-art offline algorithms (Smith and Simmons 2004; Pineau et al. 2006; Shani et al. 2007; Bonet and Geffner 2009). Here, we only use SARSOP as a representative offline algorithm to compare with our online algorithms.

In addition, published results on the partially observable Monte-Carlo planning (POMCP) algorithm (Silver and Veness 2010) for some RockSample problems are also presented. POMCP is a promising online Monte-Carlo algorithm, especially in large-scale POMDP problems. From these data, we can see FHHOP exceeds POMCP in terms of both overall running time and qualities of generated policies on these RockSample problems.

## 5 Conclusion and Future Work

This paper presents FHHOP, a novel factored hybrid heuristic online planning algorithm for large POMDPs. To the best of our knowledge, FHHOP is the most efficient online heuristic search algorithm, on the whole, yet proposed amongst existing fully implemented domain-independent online POMDP solvers. A major contribution of this algorithm is a new hybrid heuristic search strategy that takes full advantage of both lower and upper bounds on the optimal value function. Its foundation stone is a clever way of constructing a heuristic function using the lower bound. A minor contribution of this algorithm is to integrate MOMDP, a recently developed factored state representation, with a cutting-edge online algorithm and reveal its efficacy empirically. Experimental results reported here provide a comprehensive picture of FHHOP and state-of-the-art online and offline approaches on five popular benchmark domains.

There are some interesting directions for future work. First, we noticed some efficient structure learning algorithms had been proposed in factored state MDPs (Li et al. 2008; Diuk et al. 2009). An interesting research direction is to extend these existing algorithms to automatically learn the mixed observation structure in POMDPs. Second, we believe it should be possible to use other cutting-edge data structures, such as algebraic decision diagrams (Poupart 2005) or topology (Brunskill and Russell 2010), in FHHOP to succinctly represent more complex structures. Third, reparameterization (Ong et al. 2009) may give POMDPs without the mixed observation structure a chance to benefit from the MOMDP representation. Fourth, we could leverage the graph structure instead of the AND/OR tree in FHHOP to avoid redundant computations on duplicate belief states. Finally, we would also like to investigate further hybrid heuristic strategies for better performance.

**Acknowledgments**

We thank Michael L. Littman, David Hsu, Stéphane Ross, Trey Smith and anonymous reviewers for their thoughtful suggestions. This work was supported in part by the China Scholarship Council, the National Science Foundation of China under Grant Nos. 60745002, 61175057, 61105039, and the National High-Tech Research and Development Plan of China under Grant No. 2008AA01Z150.